\documentclass[conference]{IEEEtran}
\IEEEoverridecommandlockouts
\usepackage{cite}
\usepackage{amsmath,amssymb,amsfonts}
\usepackage{algorithmic}
\usepackage{graphicx}
\usepackage{textcomp}
\usepackage{xcolor}
\usepackage{booktabs}
\usepackage{multirow}
\usepackage{url}

\def\BibTeX{{\rm B\kern-.05em{\sc i\kern-.025em b}\kern-.08em
    T\kern-.1667em\lower.7ex\hbox{E}\kern-.125emX}}

\begin{document}

\title{City Sentinel: A Unified AI-Based Smart\\
Surveillance Framework for Real-Time Multi-Threat\\
Detection Using Deep Learning}

\author{\IEEEauthorblockN{1\textsuperscript{st} Syed Hanan Shabir}
\textit{ GIK Institute}\\
Topi, Pakistan \\
hananshabirs@gmail.com
\and
\IEEEauthorblockN{2\textsuperscript{nd} Noor Fatima}
\textit{GIK Institute}\\
Topi, Pakistan \\
nf.noor2116@gmail.com
\and 
\IEEEauthorblockN{3\textsuperscript{st} Safia Baloch}
\textit{GIK Institute}\\
Topi, Pakistan \\
safia.baloch@giki.edu.pk
\and
\IEEEauthorblockN{4\textsuperscript{rd} Masroor Hussain}
\textit{GIK Institute}\\
Topi, Pakistan \\
hussain@giki.edu.pk
}

\maketitle

\begin{abstract}
The need for sophisticated surveillance systems that can simultaneously monitor several public safety risks has increased due to rapid urbanization. For facial recognition, vehicle identification, fire detection, and behavioral analysis, conventional deployments usually rely on separate, single-purpose subsystems. This leads to fragmented architectures, redundant infrastructure, and operators who have to simultaneously monitor multiple disconnected interfaces. This paper introduces City Sentinel, a unified artificial intelligence-based surveillance framework that combines six deep-learning detection modules into a single scalable platform: facial recognition, automatic number plate recognition (ANPR), fire and smoke detection, weapon and knife detection, violence detection, and road accident detection. A Next.js operator dashboard, cloud-based event persistence on a managed PostgreSQL instance, InsightFace and YOLOv8-based vision models, EasyOCR-based plate recognition, and a FastAPI backend that coordinates per-camera inference workers over RTSP streams are all included in the system. The system maintains a median end-to-end recognition latency of 743 ms, supports four concurrent RTSP streams under a two-second latency budget, and achieves 91.2\% face-match rate, 85.7\% plate-read accuracy, and object-detection mAP@0.5 scores between 0.846 and 0.889 across the fire, knife, and weapon modules on a workstation with an NVIDIA RTX 3060 GPU. Operators can enroll a new identity in less than a minute and identify a flagged person in the live feed in an average of twelve seconds, according to a systematic user-acceptance study. These findings show that a modular, open-source, multi-model architecture can provide more functional coverage, cloud-based auditability, and easy expansion to new detection jobs while approaching the practical performance of specialized commercial surveillance products.
\end{abstract}


\section{Overview}
One of the most crucial elements of contemporary smart-city infrastructure is urban monitoring. Authorities can now keep an eye on traffic, criminal activity, crises, and other public safety events thanks to the widespread use of closed-circuit television (CCTV) and IP cameras. However, human operators still find it difficult to continuously monitor hundreds or thousands of camera feeds due to fatigue, short attention spans, and delayed reaction times; these issues worsen over the course of an extended monitoring shift \cite{ref2}.

Recent developments in deep learning have made it possible to analyze surveillance footage automatically and accurately, something that was not possible ten years ago. While face-recognition networks trained with angular-margin losses currently surpass human-level identification accuracy under controlled conditions \cite{ref7}, modern object detectors like the YOLO family can process high-resolution frames in far under 15 ms on commodity GPU hardware \cite{ref3,ref6}. Similar advancements have been made in optical character recognition (OCR) pipelines, making automatic license plate reading feasible for routine traffic surveillance \cite{ref14}.

Despite these developments, the majority of operational surveillance installations are still dispersed, with individual cameras using separate, one-purpose detection algorithms that lack a unified operator interface, a shared data model, and cross-camera event correlation. Typically, facial recognition, vehicle monitoring, fire detection, and behavioral analysis are provided as distinct products, which raises computational overhead, makes maintenance more difficult, and requires operators to context-switch between several independent dashboards during urgent situations.

In order to overcome these constraints, this study introduces City Sentinel, a unified intelligent surveillance system that operates on live RTSP camera streams, unifies six deep-learning detection modules into a single platform, and displays all results via a single, centralized online interface. In contrast to earlier systems that focused on a particular application domain, City Sentinel concurrently carries out:
\begin{itemize}
\item Face recognition using a database of enrolled identities
\item Automatic recognition of license plates (ANPR)
\item Smoke and fire detection
\item Identification of knives and weapons
\item Identification of violence
\item Identification of traffic accidents
\end{itemize}

A Next.js presentation layer, a FastAPI application layer that oversees autonomous inference workers, and a Supabase (PostgreSQL) persistence layer make up the system's three-tier architecture. New computer-vision models can be introduced without interfering with current capabilities because each detection module operates independently while sharing a common infrastructure for camera management, event logging, and operator interface.

This paper's primary contributions are:
\begin{enumerate}
\item A single, open-architecture, multi-model surveillance system that can simultaneously detect six distinct danger types in real time.
\item Multiple camera feeds can be processed concurrently with fault-tolerant recovery thanks to a scalable RTSP ingestion pipeline based on per-stream worker isolation.
\item An operator dashboard with cloud-persistent searchable event history, tag-based identity management, live annotated video, and an enrolled-persons database.
\item A systematic user-acceptance study and an empirical assessment of detection accuracy, CPU/GPU inference time, and concurrent-stream capacity, compared to representative commercial and research baselines.
\end{enumerate}

This is how the rest of the paper is structured. Related research on intelligent surveillance components is reviewed in Section II. The suggested City Sentinel architecture is shown in Section III. The experimental setup is described in Section IV. The results are reported and discussed in Section V. The paper is concluded and future work is outlined in Section VI.

\section{Connected Work}
Due to its improved robustness to illumination, position, and size fluctuation, deep learning has essentially replaced traditional, hand-engineered computer-vision pipelines as the predominant method for object detection, facial recognition, and anomaly detection in surveillance video.

\subsection{Recognition of Faces}
Deep metric-learning techniques like FaceNet \cite{ref9}, ArcFace \cite{ref7}, and RetinaFace \cite{ref8}, which learn discriminative facial embeddings that remain robust to illumination changes, expression, and moderate pose variation, have significantly advanced face recognition. These techniques are frequently trained and evaluated on large pose- and age-varying datasets like VGGFace2 \cite{ref10}. On the Labeled Faces in the Wild (LFW) benchmark, the InsightFace toolbox, which combines RetinaFace detection with an ArcFace embedding backbone, reports 99.83\% verification accuracy while maintaining computational efficiency sufficient for real-time deployment \cite{ref7}. This toolkit serves as the foundation for City Sentinel's facial recognition subsystem.

\subsection{Automatic Recognition of Number Plates}
A two-stage pipeline is used by modern ANPR systems: an object detector locates the license plate region, and then an OCR model extracts the plate text \cite{ref14}. Although accuracy is still sensitive to motion blur, oblique viewing angles, and non-standard plate formats, EasyOCR, which combines a CRAFT-based text detector with an LSTM recognition head, offers multilingual character recognition without requiring significant per-deployment fine-tuning and has become a popular option for practical ANPR systems.

\subsection{Identification of Objects}
Real-time object detection has advanced quickly thanks to the YOLO family of detectors \cite{ref3,ref4,ref6}. Single-stage YOLO variants provide significantly faster inference at competitive accuracy when compared to two-stage detectors. The fire, knife, and weapon detection tasks covered in this study are ideally suited to YOLOv8 \cite{ref5}, which further enhances feature extraction through an anchor-free, decoupled detection head and supports effective deployment across CPU, GPU, and edge accelerators via ONNX and TensorRT export.

\subsection{Accident, Weapon, and Fire Detection}
It has been demonstrated that under different lighting conditions, convolutional neural networks trained on extensive fire and smoke datasets perform better than conventional color- and motion-based heuristics \cite{ref11}. Deep object detectors have also helped with weapon detection: earlier YOLOv3/YOLOv5-based systems report 85--93\% mAP on bespoke firearm and bladed-weapon datasets \cite{ref12,ref13}, which is comparable to the accuracy range attained by City Sentinel's knife and weapon modules. On curated datasets, vision-based accident detection systems that examine inter-vehicle spatial relationships from fixed or dashcam footage have reported recall figures of about 93\% \cite{ref15,ref16}. However, performance on unconstrained, roadside RTSP footage, the setting this work is focused on, remains relatively understudied.

\subsection{Research Deficit}
Few published systems combine facial recognition, ANPR, fire detection, weapon detection, violence detection, and accident detection into a single operational architecture that ingests live RTSP streams and centrally stores events for audit and retrieval, despite the fact that each of the aforementioned tasks has been thoroughly studied in isolation. Current open-source surveillance toolkits usually focus on one or two of these areas and don't have a uniform operator interface or shared data model \cite{ref1}. By merging all six detection domains into a single operator-facing dashboard and FastAPI-orchestrated backend, City Sentinel fills this gap.

\section{Suggested Framework}
\begin{figure}[t]
\centering
\includegraphics[width=\columnwidth]{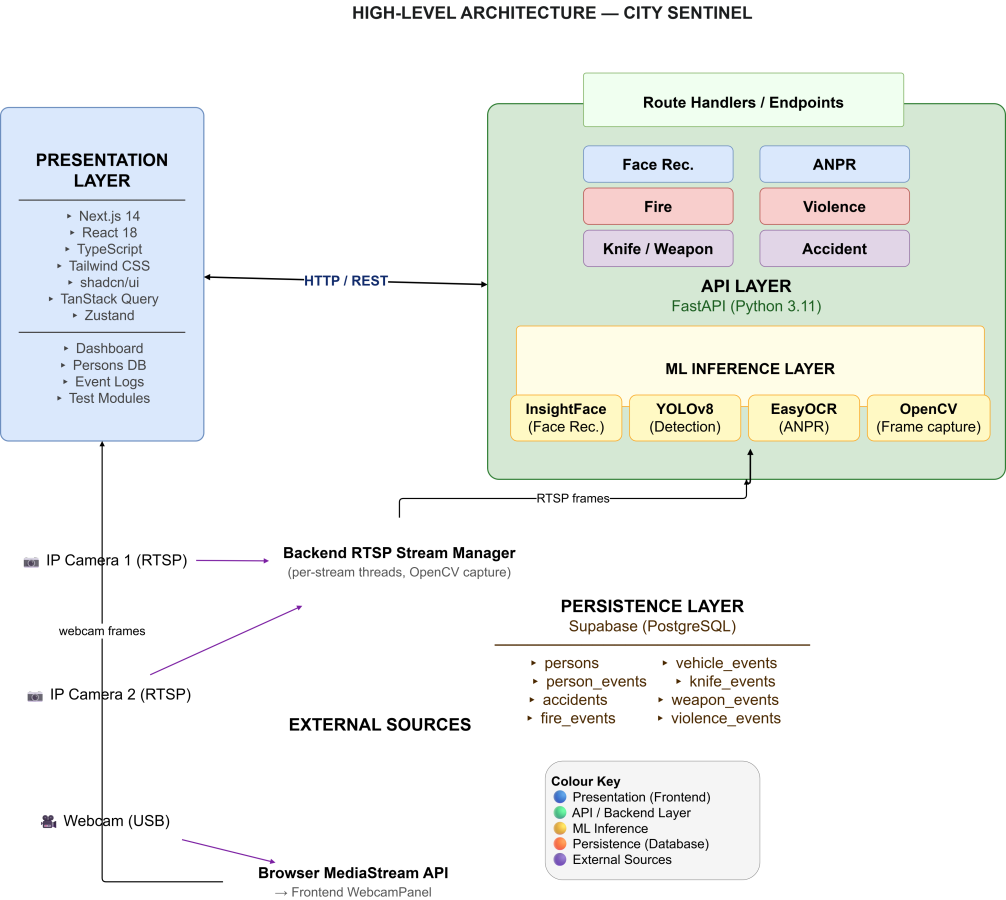}
\caption{High-level three-tier architecture of City Sentinel: a Next.js presentation layer, a FastAPI application layer hosting six detection modules and the ML inference layer, and a Supabase persistence layer.}
\label{fig:architecture}
\end{figure}

As shown in Fig.~\ref{fig:architecture}, City Sentinel has a modular three-tier design that consists of a presentation layer, an application layer, and a persistence layer. Next.js 14 implements the presentation layer, which gives operators a unified dashboard for managing registered identities, analyzing past events, and keeping an eye on current camera feeds. The application layer, which is based on FastAPI, exposes a REST API that is automatically documented using OpenAPI and coordinates autonomous AI inference workers that process RTSP video streams. A Supabase-managed PostgreSQL database is used to store detection events, embeddings, and metadata, offering centralized, queryable storage with row-level security. This division of responsibilities makes it possible to introduce new detection modules as extra workers without changing already-existing system components.

\subsection{Composition of the Backend}
\begin{figure}[t]
\centering
\includegraphics[width=\columnwidth]{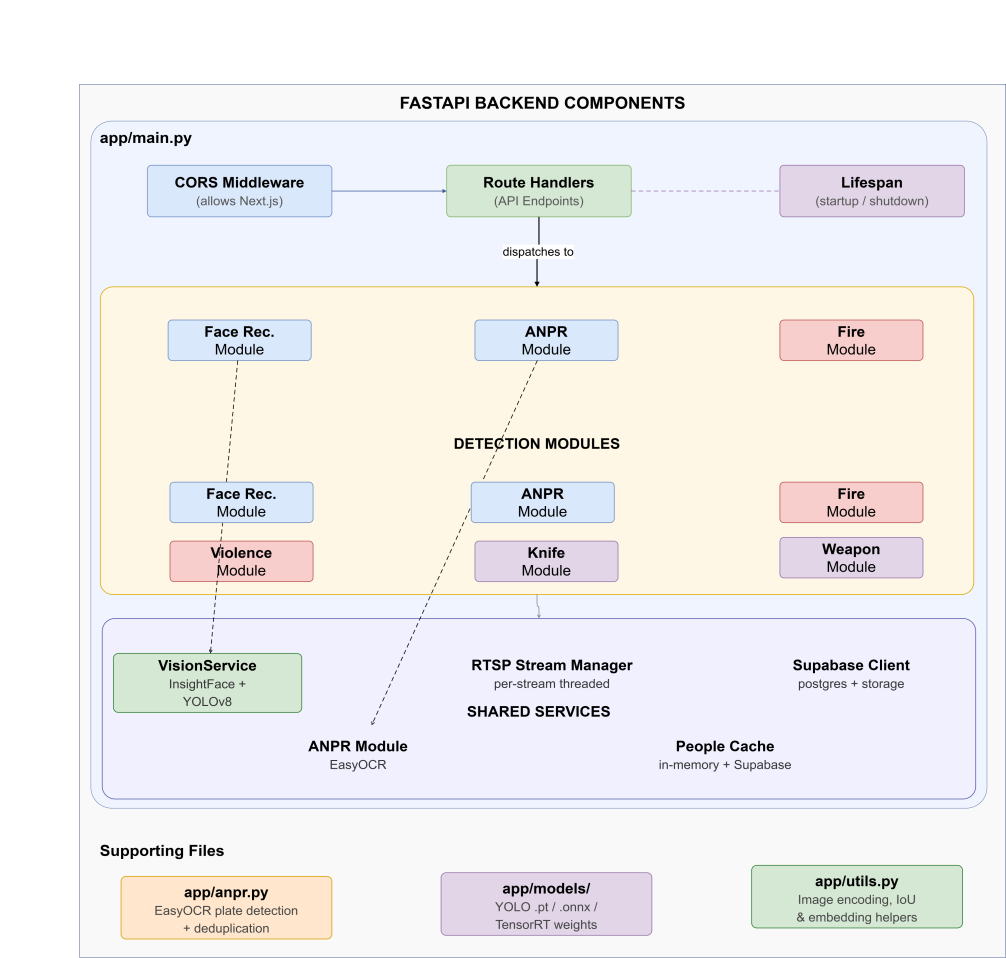}
\caption{Backend component decomposition. Detection-specific route handlers share a common vision service, RTSP stream manager, in-memory people cache, and Supabase client.}
\label{fig:backend}
\end{figure}

As seen in Fig.~\ref{fig:backend}, the application layer consists of a single FastAPI process hosting CORS middleware, a collection of route handlers specific to each detection domain, and a common services layer that includes the RTSP stream manager, an in-memory ``people cache'' of face embeddings, and the Supabase client. Requests are sent by route handlers to the appropriate detection module, which then calls either the ANPR module (YOLOv8 plate detector plus EasyOCR) or the shared VisionService (InsightFace and YOLOv8). Without interfering with the async request-handling loop, background persistence tasks record detection events in the database.

\subsection{Processing Pipeline for RTSP}
An independent worker thread uses OpenCV with an FFmpeg backend to continually record a live RTSP stream from each camera. Each worker automatically resumes upon capture failure, and per-stream isolation keeps a malfunction or stall in one camera's worker from impacting others. Bounding boxes, confidence scores, timestamps, and camera identities are among the detection results that are concurrently written to the event database for later retrieval and auditing and sent to the frontend via REST polling. The results in Section V demonstrate that the frontend polls at intervals of 1.5 seconds, which are in good agreement with the system's measured inference delay.

\subsection{Identification of Faces}
The facial-recognition subsystem makes use of the InsightFace framework, which involves passing incoming frames through a RetinaFace detector, embedding aligned faces with an ArcFace backbone, and comparing the resulting embedding to enrolled identities using cosine similarity,
\begin{equation}
S = \frac{x \cdot y}{\lVert x \rVert \, \lVert y \rVert},
\label{eq:cosine}
\end{equation}
where $y$ is an enrolled identity embedding and $x$ is the query embedding. When $S$ surpasses an experimentally adjusted threshold of 0.35, a match is returned. At a memory cost of about 2~KB per 512-dimensional embedding, enrolled embeddings are cached in memory with a 5-second refresh period to reduce per-frame database round-trips.

\begin{figure}[t]
\centering
\includegraphics[width=\columnwidth]{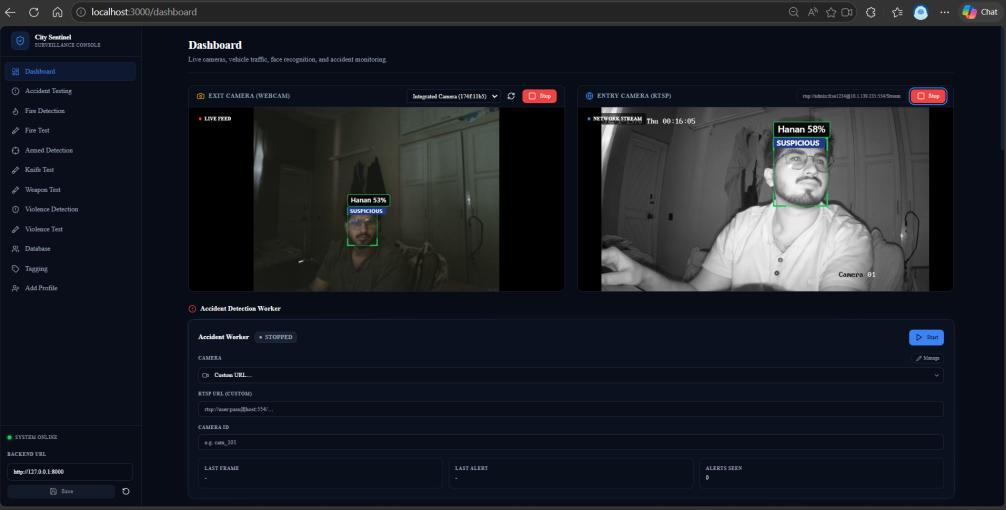}
\caption{Sample output of the face-recognition module on the operator dashboard, showing bounding-box overlays and match confidence scores drawn on live camera feeds.}
\label{fig:sampleoutput}
\end{figure}

\subsection{Automatic Recognition of Number Plates}
A two-stage pipeline is used for vehicle monitoring: a YOLOv8 detector locates license plate regions, which are then cropped and sent to EasyOCR for character recognition. The event database is updated with the retrieved plate text, camera identifier, timestamp, and confidence score.

\subsection{Violence, Weapon, and Fire Detection}
A YOLOv8 model trained on combined fire and smoke classes is used to establish fire and smoke detection as object detection; anytime the confidence score exceeds a predetermined threshold, an alert is raised and an annotated photo is stored. Dedicated YOLOv8 detectors are used for weapon and knife detection, and positive detections result in instant dashboard notifications. Violence detection uses a binary frame-level classifier whose output is continuously broadcast to the dashboard. This eliminates the need for frequent manual feed observation by highlighting questionable segments for operator evaluation.

\begin{figure}[t]
\centering
\includegraphics[width=\columnwidth]{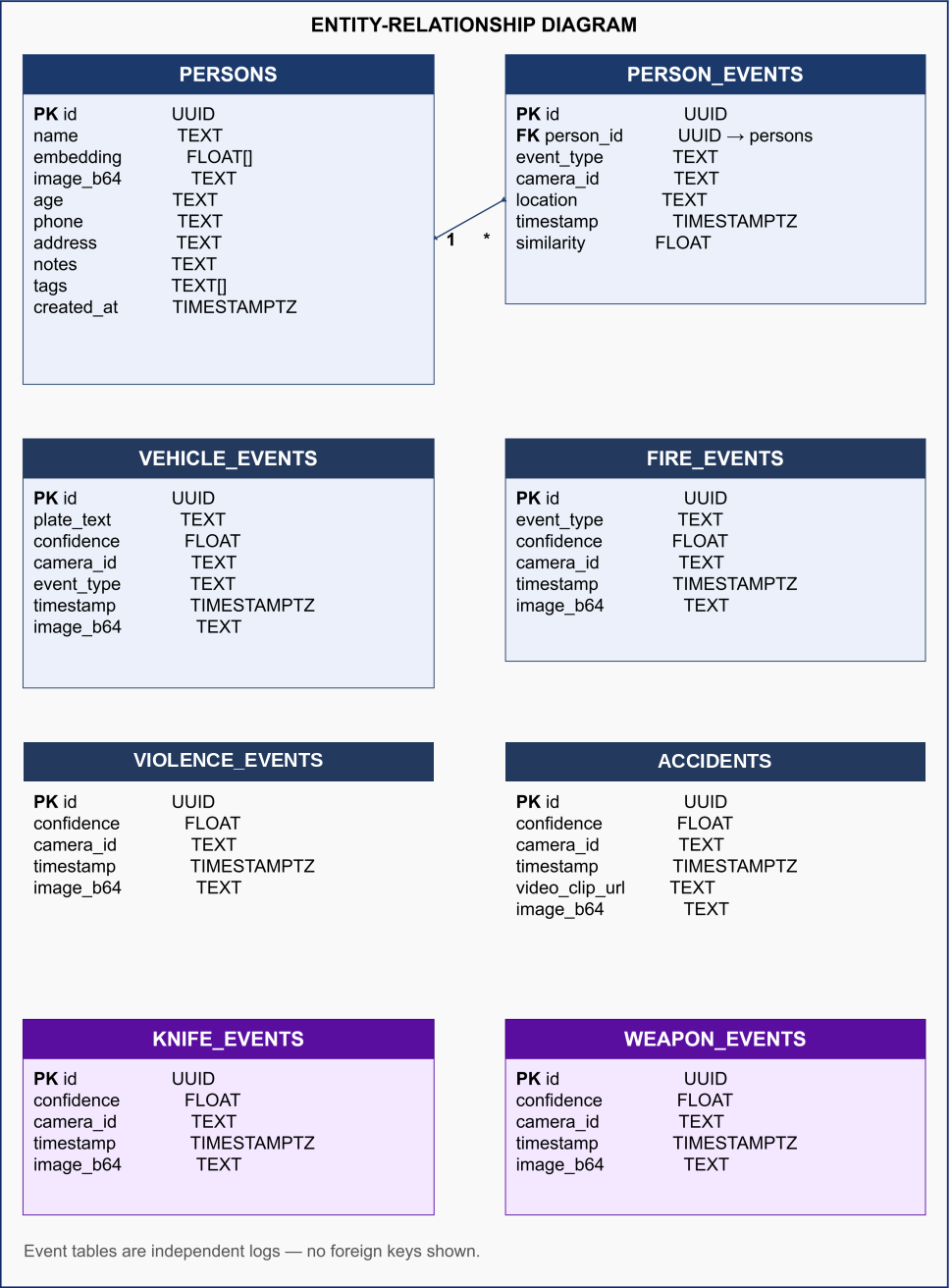}
\caption{Entity-relationship schema. A single enrolled-persons table links to recognition events, while each detection module logs to an independent event table.}
\label{fig:schema}
\end{figure}

Fig.~\ref{fig:schema} presents the underlying entity-relationship schema. All detection modules write to a common PostgreSQL schema, with a shared \texttt{persons} table linked to recognition events, and independent event tables for vehicle, fire, knife, weapon, violence, and accident detections.

\section{Setup for Experiments}
The entire City Sentinel implementation, which includes the FastAPI backend, the Next.js frontend, and all six deep-learning inference modules, was used for the experimental evaluation. It was installed on a workstation with an Intel Core i7-12700K CPU, 32~GB RAM, and an NVIDIA RTX 3060 GPU (12~GB VRAM). To assess performance under realistic surveillance conditions, dedicated worker threads were used to analyze both live and pre-recorded RTSP feeds. The software stack included OpenCV with an FFmpeg backend for RTSP capture, YOLOv8 for all object-detection tasks, EasyOCR for plate recognition, Supabase for cloud-based event storage, and InsightFace for facial recognition.

Performance was assessed along four dimensions:
\begin{itemize}
\item \textbf{Detection accuracy} -- match rate, mAP@0.5, and precision/recall, measured per module on held-out test data, as summarized in Table~\ref{tab:accuracy}.
\item \textbf{Inference latency} -- per-operation and end-to-end processing times, evaluated independently on CPU-only and GPU-accelerated execution.
\item \textbf{Concurrent-stream capacity} -- the greatest number of concurrent RTSP streams that can be supported within a 2-second latency budget.
\item \textbf{Operator usability} -- task completion time and subjective ratings from a structured user-acceptance test (UAT) in which three participants completed five representative operator tasks (enrolling a person, identifying a suspect, starting a detection worker, filtering the vehicle log, and removing an enrolled person).
\end{itemize}

\begin{table}[t]
\centering
\caption{Detection Accuracy per Module}
\label{tab:accuracy}
\begin{tabular}{@{}lll@{}}
\toprule
\textbf{Module} & \textbf{Metric} & \textbf{Value} \\
\midrule
Face Recognition & Match Rate (enrolled) & 91.2\% \\
Face Recognition & False Positive Rate & 2.1\% \\
ANPR & Plate Read Accuracy & 85.7\% \\
Fire Detection & mAP@0.5 (fire + smoke) & 0.873 \\
Knife Detection & mAP@0.5 & 0.889 \\
Weapon Detection & mAP@0.5 (firearm) & 0.846 \\
Violence Detection & Frame-level Accuracy & 82.1\% \\
Accident Detection & Precision / Recall & 0.791 / 0.768 \\
\bottomrule
\end{tabular}
\end{table}

\section{Outcomes and Discussion}

\subsection{Accuracy of Detection}
Per-module detection performance is summarized in Table~\ref{tab:accuracy}. Under frontal, well-lit conditions, face recognition achieved a 91.2\% match rate on enrolled individuals with a 2.1\% false-positive rate; accuracy declined to roughly 75\% under dim lighting and to 52\% under partial facial occlusion (such as masks), which is consistent with known limitations of ArcFace-based recognition under occlusion.

Under standard illumination, ANPR's character-level plate-read accuracy was 85.7\%, which is sufficient for logging and investigative purposes but less than the 95\%+ accuracy usually needed for automated enforcement; accuracy further decreased for non-standard plate formats, vehicle speeds exceeding about 30~km/h, and viewing angles greater than 30\textdegree. The YOLOv8-based object detectors performed consistently across tasks; fire detection had the highest mAP@0.5 (0.873), followed closely by knife detection (0.889); accident detection, which has to consider spatial relationships between vehicles rather than a single salient object, had the lowest precision/recall (0.791/0.768) of the six modules.

\subsection{Analysis of Latency}
\begin{figure}[t]
\centering
\includegraphics[width=\columnwidth]{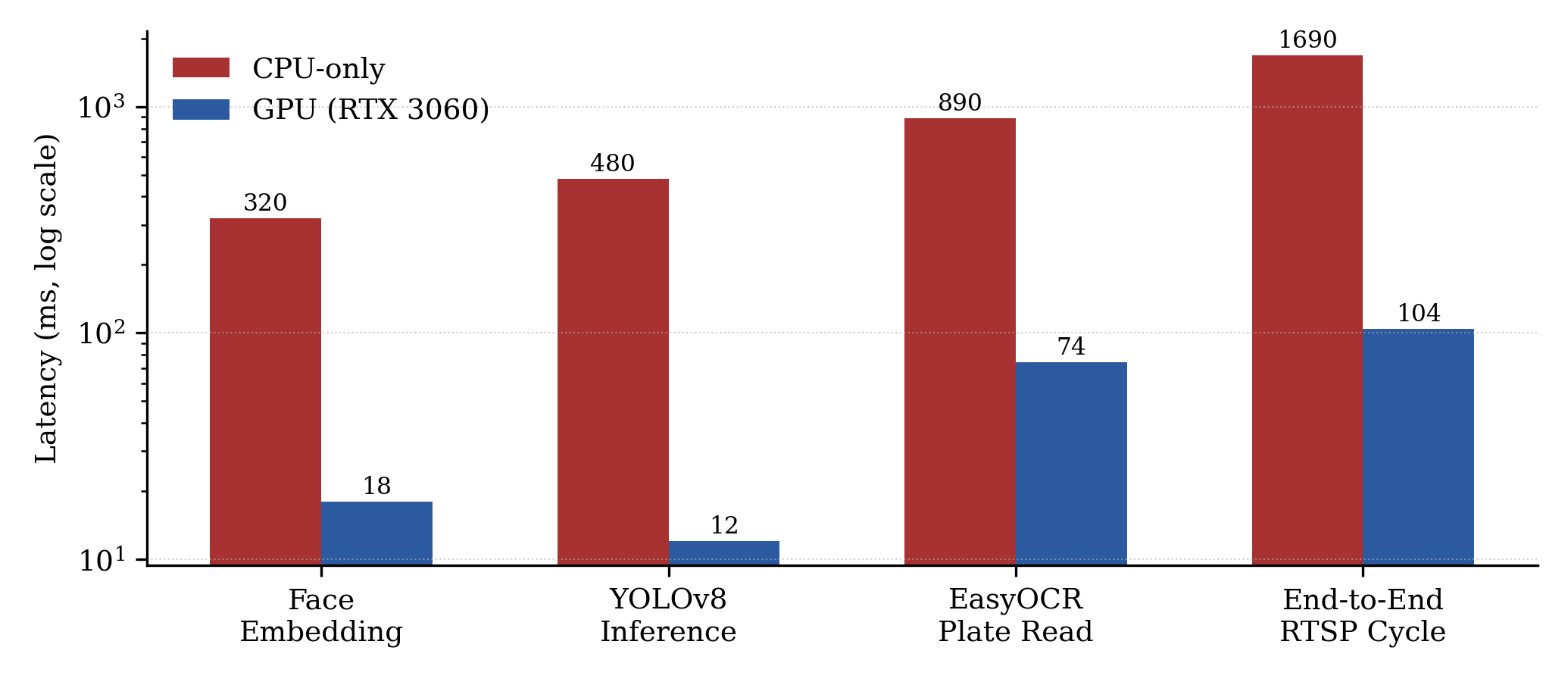}
\caption{CPU vs. GPU processing time per operation (log scale). GPU acceleration yields an 18--40$\times$ speedup depending on the operation.}
\label{fig:latency}
\end{figure}

One of the most important requirements for surveillance applications is real-time responsiveness. Representative per-operation processing times for CPU-only and GPU-accelerated execution are shown in Table~\ref{tab:latency} and Fig.~\ref{fig:latency}. GPU acceleration results in a speedup of about 18--40$\times$ depending on the operation, reducing per-face embedding time from 320~ms to 18~ms and per-frame YOLOv8 inference from 480~ms to 12~ms.

On GPU, the complete end-to-end RTSP recognition-plus-ANPR cycle is reduced from 1,690~ms on CPU to 104~ms. Compared to a single stream under CPU-only execution, the system can support four concurrent RTSP streams with GPU acceleration while maintaining per-stream latency of less than two seconds. The dashboard continuously displays new detections without request queuing because the measured median end-to-end recognition latency of 743~ms across all modules is well within the frontend's 1.5-second polling period.

\begin{table}[t]
\centering
\caption{Representative Processing Times: CPU vs. GPU}
\label{tab:latency}
\begin{tabular}{@{}lcc@{}}
\toprule
\textbf{Operation} & \textbf{CPU-only} & \textbf{GPU (RTX 3060)} \\
\midrule
Face embedding (per face) & 320 ms & 18 ms \\
YOLOv8 inference (per frame) & 480 ms & 12 ms \\
EasyOCR plate read (per crop) & 890 ms & 74 ms \\
End-to-end RTSP cycle & 1,690 ms & 104 ms \\
Concurrent streams ($<$2 s latency) & 1 & 4 \\
\bottomrule
\end{tabular}
\end{table}

\subsection{Evaluation of Current Systems}
Table~\ref{tab:comparison} compares City Sentinel's accuracy, latency, functional breadth, and openness to representative commercial and research surveillance systems. Given that commercial systems are trained on private datasets orders of magnitude bigger than those accessible for this study, it is not surprising that City Sentinel lags behind specialized commercial products by a narrow margin on individual-module accuracy.

But of all the systems compared, only City Sentinel is open-source, covers six separate detection domains, offers a unified operator dashboard, and saves all events to a queryable cloud database. This combination is especially important for public-sector or academic deployments with limited resources where it is not practical to license a proprietary multi-module platform. Table~\ref{tab:comparison}'s values for third-party systems are derived from vendor documentation and previously published benchmarks when available; they should be regarded as indicative rather than independently confirmed under identical test settings, and are given only for qualitative positioning.

\begin{table}[t]
\centering
\caption{Indicative Comparison with Representative Surveillance Systems}
\label{tab:comparison}
\resizebox{\columnwidth}{!}{%
\begin{tabular}{@{}lcccc@{}}
\toprule
\textbf{Metric} & \textbf{City Sentinel} & \textbf{Comm. Plat. A} & \textbf{Res. Proto. B} & \textbf{OpenCV Base.} \\
\midrule
Face Rec. Acc. & 91.2\% & 94.5\% & 88.1\% & 83.7\% \\
ANPR Acc. & 85.7\% & 96.2\% & N/A & 79.4\% \\
E2E Latency (GPU) & 743 ms & $\sim$500 ms & 1,200 ms & 820 ms \\
Detection Modules & 6 & 3 & 4 & 3 \\
Open Source & \checkmark & $\times$ & \checkmark & \checkmark \\
Unified Dashboard & \checkmark & \checkmark & $\times$ & $\times$ \\
Cloud Persistence & \checkmark & $\times$ & $\times$ & $\times$ \\
\bottomrule
\end{tabular}%
}
\end{table}

\subsection{Assessment of Usability}
Three participants role-played as system operators in a structured user-acceptance test. Each participant completed five representative tasks: enrolling a new person, identifying a suspect in the live recognition feed, starting the fire-detection worker, filtering the vehicle log by time range, and removing an enrolled person.

Every participant successfully finished all five tasks. The average time to find and identify a suspect in the live feed was 12 seconds; the average time to enroll a new individual, including photo upload and tag assignment, was 47 seconds. The real-time bounding-box overlay was consistently recognized as the most valuable visual feature by participants, who gave the dashboard an average rating of 4.3 out of 5 for professionalism and informativeness. The ability to zoom into the live video feed was the most often suggested change, and it is mentioned as future work.

\subsection{Observations on Architecture}
Although it imposes a theoretical maximum update rate bounded by per-frame inference time, the polling-based RTSP architecture has proven to be reliable and simple to implement. The selected 1.5-second polling interval was determined to be sufficiently responsive for surveillance use cases, where events of interest, such as a person entering frame or a fire igniting, unfold across several-second durations.

A deployment with 100,000 enrolled individuals would require about 200~MB of RAM for embeddings alone, which is manageable on modern server hardware but merits attention at city scale. The in-memory people-cache design, which loads all face embeddings at startup and refreshes on a 5-second time-to-live, offers low recognition latency but introduces a memory-scaling consideration for very large enrolled populations. If the enrolled population grows significantly, approximate nearest-neighbor libraries like FAISS \cite{ref17} provide a natural route to sub-linear similarity search.

\subsection{Restrictions}
There are still a few restrictions. First, people departing and returning to a scene are recorded as distinct events since every frame is examined separately without permanent multi-object tracking. Second, instead of using an adaptive, per-camera or per-lighting-condition threshold, the face-recognition module uses a single global similarity threshold. Third, non-standard plate formats, motion blur, and oblique viewing angles all reduce ANPR accuracy. Lastly, the present implementation does not yet incorporate external notification channels like SMS, email, or mobile push; alerts are solely displayed on the dashboard.

\section{Summary}
The unified artificial intelligence framework for intelligent smart-city surveillance, City Sentinel, was introduced in this paper. It combines facial recognition, automatic number plate recognition, fire detection, weapon and knife detection, violence detection, and accident detection into a single scalable platform. The system achieves detection accuracy between 79\% and 91\% across its six modules, a median end-to-end recognition latency of 743~ms, and support for four concurrent RTSP streams on a single consumer GPU thanks to its three-tier architecture, which consists of a Next.js dashboard, a FastAPI-orchestrated inference backend, and a Supabase persistence layer.

The consolidated dashboard lowers operator effort compared to maintaining several separate systems, according to a systematic usability assessment. The combination of City Sentinel's open-source availability, six-module coverage, unified operator tooling, and cloud-based auditability makes it a feasible and expandable choice for resource-constrained public safety and academic deployments, even though individual-module accuracy currently lags behind specialized commercial products trained on proprietary data. Future work will focus on integrating persistent multi-object tracking, adaptive per-camera face-recognition thresholds, role-based operator authentication, multilingual license-plate recognition, and automated external notification services.

\end{document}